\documentclass[10pt,twocolumn]{article}

\usepackage[margin=0.75in]{geometry}
\usepackage{times}
\usepackage{microtype}
\usepackage{graphicx}
\usepackage{booktabs}
\usepackage{amsmath,amssymb}
\usepackage{hyperref}
\usepackage{url}
\usepackage{caption}
\usepackage{listings}

\hypersetup{
  colorlinks=true,
  linkcolor=blue,
  citecolor=blue,
  urlcolor=blue
}

\title{SLO-Conditioned Action Routing for Retrieval-Augmented Generation:\\Objective Ablation and Failure Modes}
\author{Bharath Nunepalli\\\href{https://bh3r1th.medium.com}{\nolinkurl{https://bh3r1th.medium.com}}\\\href{https://github.com/bh3r1th/rl-rag-slo-controller}{\nolinkurl{https://github.com/bh3r1th/rl-rag-slo-controller}}}
\date{}

\begin{document}

\twocolumn[
\maketitle

\begin{abstract}
Retrieval-augmented generation (RAG) introduces a practical control problem: retrieval depth and generation behavior must be chosen per query to satisfy service-level objectives (SLOs) such as cost, refusal rate, and hallucination risk.
This work models per-query control as a small discrete action: choose a retrieval depth $k\in\{2,5,10\}$ and a generation mode (\textit{guarded} vs.\ \textit{auto}), or refuse.
An offline logged dataset is constructed from SQuAD~2.0 by executing each action and recording accuracy, token cost, hallucination/refusal indicators, and an SLO-weighted reward.
Two simple policy-learning objectives are evaluated: supervised classification of the per-state best action (Argmax-CE) and a reward-weighted variant (Argmax-CE-WT).
Across the evaluated settings, a strong fixed baseline (low $k$, guarded prompting) performs competitively; learned policies mainly provide additional cost savings under a quality-focused SLO and can exhibit \textit{refusal collapse} under a cheap SLO when refusal is heavily rewarded.
The contribution is a reproducible case study of SLO-aware control for RAG pipelines, emphasizing failure modes and reporting conventions rather than proposing a new retriever or language model.
\end{abstract}
\vspace{0.25in}
]

\section{Introduction}
RAG systems are rarely a single fixed algorithm in practice. Engineers choose retrieval depth, prompt structure, and safety constraints, and each knob trades off cost, latency, and answer quality.
A configuration that is acceptable for a short factual question can be wasteful (or unsafe) for an ambiguous or unanswerable one.
In other words, \emph{RAG is a control problem}: the system must select an operating point per query to satisfy service-level objectives (SLOs).

This work studies a minimal but concrete version of that problem.
For each query, the controller chooses among a small discrete set of actions: retrieve top-$k$ passages with $k\in\{2,5,10\}$ and answer in either a \textit{guarded} mode (answer only from retrieved context, otherwise refuse) or an \textit{auto} mode (standard answering), or refuse immediately.
Each action is evaluated with an LLM and logged into an offline dataset; different SLO profiles assign different weights to accuracy, token cost, hallucination risk, and refusal behavior.

The goal is not to claim state-of-the-art QA performance.
Instead, the goal is to provide a reproducible case study of SLO-conditioned routing for RAG, including an explicit treatment of a common failure mode: \emph{refusal collapse} when a ``cheap'' reward over-incentivizes refusal.
The main contributions are:
\begin{itemize}
  \item a small open-source testbed for SLO-aware RAG control with logged offline feedback on SQuAD~2.0;
  \item two simple offline policy-learning baselines (Argmax-CE and Argmax-CE-WT) and a reporting template centered on cost--quality trade-offs; and
  \item an empirical analysis showing when learned policies help and when strong fixed actions \mbox{dominate}, plus a practical mitigation for refusal collapse.
\end{itemize}

\section{Related work}
\textbf{Retrieval-augmented generation and learned retrieval.}
RAG \cite{lewis2020rag} and dense retrieval methods such as DPR \cite{karpukhin2020dpr} and REALM \cite{guu2020realm} established a standard recipe for grounding generation on retrieved evidence. 
Recent surveys summarize a rapidly growing design space spanning retrievers, chunking, reranking, and context composition \cite{gao2024ragsurvey}. 
Several recent systems introduce \emph{adaptive} retrieval or self-reflection mechanisms to decide when and what to retrieve (e.g., Self-RAG) \cite{asai2023selfrag}. 
This work is narrower: it does not alter the retriever or the LLM, and it focuses on a small discrete action space motivated by operational SLO trade-offs.

\textbf{RL, bandits, and learning from logged rewards.}
Choosing an action per query with a scalar outcome naturally connects to contextual bandits and off-policy learning from logged feedback. 
Counterfactual risk minimization (CRM) formalizes learning policies from logged bandit data \cite{swaminathan2015crm}, and doubly robust estimators reduce variance in off-policy evaluation and learning \cite{dudik2011dr}. 
Offline RL surveys discuss broader settings with sequential decision making and distribution shift \cite{levine2020offline}. 
This work adopts a pragmatic approach: it precomputes per-action rewards for each query and trains lightweight policies by (i) reward-weighted learning and (ii) supervised best-action classification; it discusses limitations and future work on counterfactual estimators in Section~\ref{sec:limitations}.

\textbf{SLOs for AI systems.}
SLOs and error budgets are a standard framework in site reliability engineering \cite{beyer2016sre}. 
In LLM systems, token cost and latency are first-class constraints; the evaluation explicitly reports cost and quality metrics per SLO profile to make the trade-offs concrete.

\textbf{LLM routing and frugality.}
Several recent systems route each request to a cheaper or more capable model (or to a tiered cascade) to control cost while maintaining quality, e.g., FrugalGPT \cite{frugalgpt2023} and RouteLLM \cite{routellm2024}.
The present work studies a complementary routing problem: the language model is held fixed while the controller selects among retrieval depth, prompting mode, and refusal as operational actions under an explicit SLO reward.

\section{Problem setup}
\subsection{RAG action space}
For each question $q$, the controller chooses an action $a \in \mathcal{A}$ from a small, fixed set:
\begin{itemize}
  \item \textbf{Action 0:} retrieve $k{=}2$ passages, \emph{guarded} generation
  \item \textbf{Action 1:} retrieve $k{=}5$ passages, \emph{guarded} generation
  \item \textbf{Action 2:} retrieve $k{=}10$ passages, \emph{guarded} generation
  \item \textbf{Action 3:} retrieve $k{=}5$ passages, \emph{auto} (less restrictive) generation
  \item \textbf{Action 4:} \emph{refuse} (no retrieval; return an abstention)
\end{itemize}
\noindent\textbf{Refusal semantics.} Action~4 is a \emph{pre-retrieval} abstention decision made from the query alone (the system returns a short refusal without fetching passages). In contrast, \emph{guarded} generation can also produce a \emph{post-retrieval} refusal when retrieved evidence is insufficient.

Guarded generation uses a stricter prompt that encourages quoting from retrieved text and refusing when evidence is insufficient; auto generation is more permissive.

\subsection{SLO profiles and reward}
An SLO profile is represented by a vector of weights that trades off answer quality against operational cost and risk.
For each $(q,a)$ the RAG pipeline is executed and scalar reward is computed
\begin{equation}
  r(q,a) = w_{\text{acc}}\cdot \text{Acc} - w_{\text{cost}}\cdot \text{Cost} - w_{\text{hall}}\cdot \text{Hall} + w_{\text{ref}}\cdot \text{Ref},
\end{equation}
where \text{Acc} is answer correctness, \text{Cost} is token cost, \text{Hall} indicates hallucination/incorrect answering behavior, and \text{Ref} captures correct refusals (and penalizes incorrect refusals).
Two profiles are considered:
\textbf{quality\_first} (heavier weight on correctness/hallucination avoidance) and \textbf{cheap} (heavier weight on token cost).

\subsection{State representation}
The controller observes a state vector $s(q)$ consisting of a question embedding and lightweight metadata features (e.g., length and simple uncertainty indicators computed from retrieval scores).
The policy outputs a categorical distribution $\pi_\theta(a\mid s)$.

\section{Methods}
\subsection{Offline log generation}
An offline replay dataset is constructed by sampling questions and, for each question, running \emph{all} actions in $\mathcal{A}$ to obtain per-action reward and metrics.
This yields tuples $(s, a, r, \text{metrics})$ for offline training and analysis.
This ``full action sweep'' is compute-expensive but simplifies offline learning because the best action under a given SLO is observable.

\subsection{Policy training objectives}
Let $a^\star(s)$ denote the empirically best action for state $s$ under the training SLO (ties broken deterministically).
Two objectives are compared:
\begin{itemize}
  \item \textbf{Argmax-CE:} supervised multi-class classification of $a^\star(s)$ using cross-entropy.
  \item \textbf{Argmax-CE-WT:} cross-entropy with per-example weighting proportional to the action-margin between the best and second-best action (favoring ``clear'' decisions).
\end{itemize}
These are not claimed as novel algorithms; they are deliberately simple objectives used to probe behavioral differences and failure modes in SLO-conditioned routing.

\subsection{Retriever, evaluation, and prompt templates}
\paragraph{Retrieval corpus and hit rate.}
The retrieval corpus consists of SQuAD~2.0 context paragraphs.
For a query, a lightweight \emph{sparse lexical retriever} ranks paragraphs and returns the top-$k$ (or $k{=}0$ for refusal). In this implementation, ranking is BM25-style bag-of-words scoring over the raw SQuAD paragraphs \cite{robertson1995okapi,robertson2009prf}; no dense retriever (e.g., DPR/embedding search) or reranker is used.
The \texttt{retrieval\_hit\_rate} metric is computed for \emph{answerable} questions only: it is the fraction of answerable examples whose retrieved set contains at least one paragraph that includes the gold answer string.\footnote{This metric is intentionally simple and favors extractive settings; it is reported to make retrieval behavior visible even when end-to-end answer accuracy is limited by the generator.}

\paragraph{Prompting modes.}
In \textit{guarded} mode, the generator is instructed to answer using only the retrieved context and to refuse if the answer is not supported.
In \textit{auto} mode, the generator answers normally given the same retrieved context.
A fixed refusal action returns a short refusal response without retrieval.
Appendix~\ref{app:prompts} lists the concrete prompt templates used by the implementation.

\section{Experiments}
\subsection{Dataset and evaluation}
SQuAD~2.0~\cite{rajpurkar2018squad2} is used, and evaluation is reported on $N{=}200$ examples from the development set.
Questions are a mixture of answerable and unanswerable cases.
This work reports:
\textbf{accuracy} (normalized exact match on answer strings),
\textbf{avg\_cost\_tokens} (prompt+completion tokens),
\textbf{hallucination\_rate} (incorrect answer given when a refusal would be appropriate),
\textbf{refusal\_rate}, and
\textbf{retrieval\_hit\_rate} (fraction of answerable questions where the gold answer string appears in the retrieved passages).

\subsection{Reproducibility and artifacts}
The full codebase for the experiments, including scripts to precompute logged action outcomes, train SLO-conditioned policies, run evaluations, and regenerate the figures, is available at \href{https://github.com/bh3r1th/rl-rag-slo-controller}{\nolinkurl{https://github.com/bh3r1th/rl-rag-slo-controller}}.
All results in this paper are produced from the public SQuAD~2.0 dataset and the released scripts; no proprietary data or employer resources are used.

\subsection{Baselines}
Learned policies are compared against:
\begin{itemize}
  \item \textbf{Fixed-$k$ baseline (action 1):} always choose action 1 ($k{=}5$, guarded).
  \item \textbf{Best fixed action:} choose the single action that maximizes average reward for the given SLO on the evaluation set.
\end{itemize}

\subsection{LLM backend}
Generation uses \texttt{gpt-4.1-nano} via the OpenAI API \cite{openai_models}.
The same backend is used for all runs to isolate routing behavior.

\section{Results}
\subsection{Strong fixed baselines and modest gains}
Table~\ref{tab:main} summarizes the key results.
Across all conditions, the \emph{best fixed action} is action~0 (retrieve $k{=}2$ with guarded generation), underscoring how strong conservative fixed policies can be.
Under \textbf{quality\_first}, Argmax-CE improves average reward from $-0.0419$ (best fixed) to $-0.0287$ and accuracy from $0.250$ to $0.275$, at higher cost (244 to 359 tokens on average).

\subsection{Behavioral shift and SLO mismatch}
\begin{figure}[t]
  \centering
  \includegraphics[width=\columnwidth]{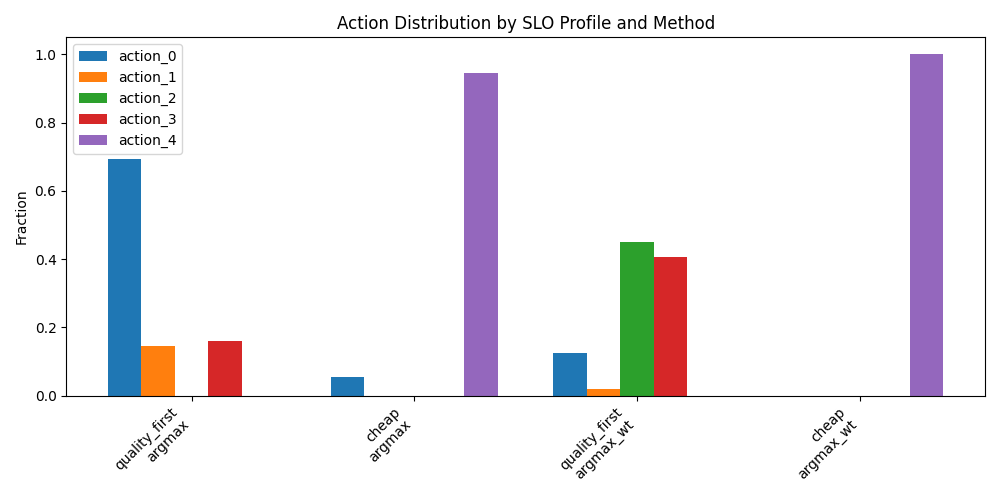}
  \caption{Action distribution of learned policies under each SLO and objective. Cost-focused settings can collapse to refusal-heavy policies.}
  \label{fig:actions}
\end{figure}

Figure~\ref{fig:actions} shows the learned action distributions.
Under \textbf{quality\_first}, Argmax-CE mixes actions, primarily selecting action 0 but sometimes choosing action 1 or action 3.
In contrast, under \textbf{cheap}, Argmax-CE collapses to refusal ($94.5\%$), yielding near-zero accuracy.

\subsection{Weighted objective instability}
Argmax-CE-WT under \textbf{quality\_first} shifts heavily toward expensive actions (actions 2 and 3), achieving higher retrieval hit-rate but \emph{worse} reward than the best fixed baseline.
This illustrates a common pitfall: weighting schemes intended to emphasize ``confident'' training examples can amplify reward-model imperfections and overfit to actions that increase retrieval coverage without improving end-task reward.

\begin{figure}[t]
  \centering
  \includegraphics[width=\columnwidth]{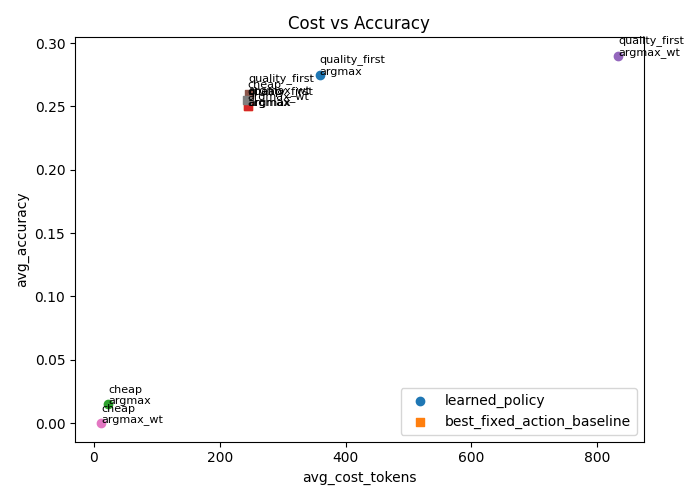}
  \caption{Average token cost vs.\ accuracy for learned policies and best fixed-action baselines.}
  \label{fig:costacc}
\end{figure}

\begin{figure}[t]
  \centering
  \includegraphics[width=\columnwidth]{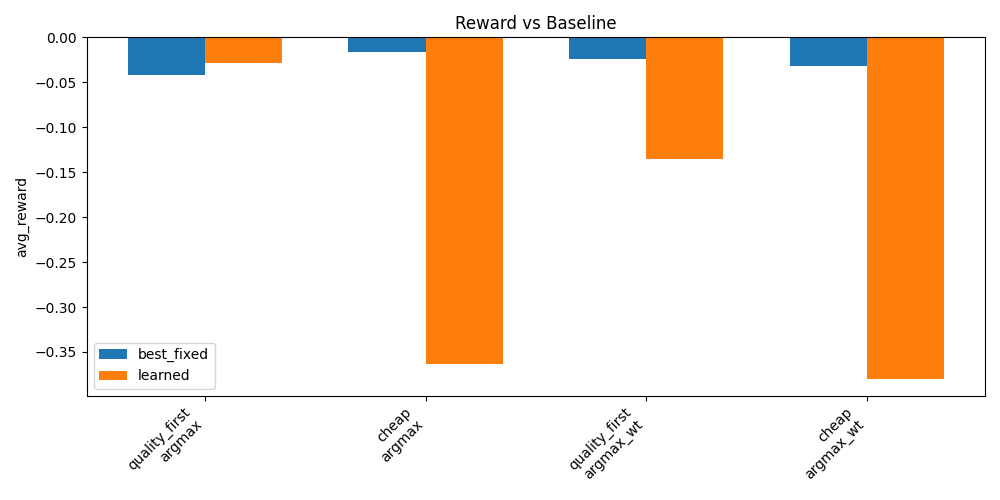}
  \caption{Average reward for best fixed actions vs.\ learned policies under each condition.}
  \label{fig:reward}
\end{figure}

\begin{table*}[t]
\centering
\caption{Key metrics on SQuAD~2.0 dev ($N{=}200$). ``Baseline'' is fixed action 1 ($k{=}5$, guarded). ``Best fixed'' is the single best action by average reward (action 0 in all cases). Retrieval hit-rate is computed over answerable questions only.}
\label{tab:main}
\small
\begin{tabular}{llrrrrrrrr}
\toprule
SLO & Method & \multicolumn{1}{c}{Acc} & \multicolumn{1}{c}{Cost} & \multicolumn{1}{c}{Reward} & \multicolumn{1}{c}{Refuse} & \multicolumn{1}{c}{Hit} &
\multicolumn{1}{c}{BestFixed Acc} & \multicolumn{1}{c}{BestFixed Cost} & \multicolumn{1}{c}{BestFixed Reward} \\
\midrule
quality\_first & Baseline (a1) & 0.300 & 609 & -0.0578 & 0.285 & 0.758 & 0.250 & 244 & -0.0419 \\
quality\_first & Argmax-CE & 0.275 & 359 & -0.0287 & 0.295 & 0.681 & 0.250 & 244 & -0.0419 \\
quality\_first & Argmax-CE-WT & 0.290 & 833 & -0.1350 & 0.185 & 0.791 & 0.260 & 246 & -0.0236 \\
\midrule
cheap & Baseline (a1) & 0.305 & 608 & -0.0521 & 0.280 & 0.758 & 0.250 & 246 & -0.0166 \\
cheap & Argmax-CE & 0.015 & 23 & -0.3638 & 0.955 & 0.033 & 0.250 & 246 & -0.0166 \\
cheap & Argmax-CE-WT & 0.000 & 11 & -0.3797 & 1.000 & 0.000 & 0.255 & 244 & -0.0323 \\
\bottomrule
\end{tabular}
\end{table*}

\section{Discussion}
\subsection{Why does refusal collapse happen?}
In offline routing, cost-heavy rewards can make ``refuse'' appear disproportionately attractive, especially when the generator is penalized for incorrect answers and the reward is measured per-query without explicit constraints on abstention.
This is a known degeneracy in selective prediction: optimizing for risk/cost without a calibrated abstention constraint often yields ``always abstain'' solutions.

\subsection{Retrieval hit-rate is necessary but not sufficient}
Argmax-CE-WT under quality\_first produces the highest retrieval hit-rate but yields lower reward than the best fixed policy.
This suggests that increasing retrieval coverage and context length can raise costs and introduce distractors, and that downstream generation quality (prompting, grounding, and refusal behavior) remains a bottleneck.

\section{Limitations}\label{sec:limitations}
This study is intentionally small and is best read as a systems-oriented case study rather than a benchmark claim.

\begin{itemize}
  \item \textbf{Sample size and statistical uncertainty.} Most reported metrics are estimated on $N{=}200$ SQuAD~2.0 dev examples per condition.
  The paper reports point estimates only; it does not provide confidence intervals, hypothesis tests, or multiple-seed analysis.
  Small differences (e.g., a few percentage points) should not be over-interpreted.
  \item \textbf{Offline evaluation and distribution shift.} Policies are trained from logged outcomes produced by a specific generator and retriever configuration.
  Any change in model, prompt, or corpus can change the reward landscape, so learned policies may not transfer without re-logging.
  \item \textbf{Simplified retrieval success metric.} \texttt{retrieval\_hit\_rate} checks whether any retrieved paragraph contains the gold answer string (answerable questions only).
  This is a coarse proxy and does not capture semantic support, multi-hop reasoning, or non-extractive settings.
  \item \textbf{Cost proxy.} Token count is used as the primary cost proxy. Real deployments also depend on latency, throughput limits, and vendor-specific pricing tiers.
\end{itemize}

\section{Conclusion}
SLO-conditioned routing is a practical control layer for RAG systems, but offline objective choice and reward design strongly shape learned behavior.
In the experiments, conservative fixed policies remain strong, quality-first routing yields modest gains, and cost-focused routing can collapse to degenerate refusal behavior.
Applied teams should report both improvements and failure modes, and treat routing as a system-design problem rather than a purely accuracy-driven benchmark.

\clearpage
\section*{Disclaimer}
This work and the accompanying code were created by the author in a personal capacity.
They are not affiliated with, endorsed by, or representative of any current or past employer.
No proprietary datasets, internal systems, or confidential resources were used; all experiments rely on publicly available datasets and public APIs.

\section*{Acknowledgments and AI assistance}
The author used AI-assisted tools (ChatGPT and OpenAI Codex) to help draft and edit parts of this manuscript and to scaffold portions of the accompanying codebase (e.g., boilerplate scripts and documentation).
All experimental design choices, code changes, results, figures, and final wording were reviewed and edited by the author, who takes full responsibility for the content.

\appendix
\section{Prompt templates}\label{app:prompts}
The implementation uses two prompting modes that share the same retrieved context but differ in safety constraints.

\subsection{Guarded mode}
\begin{lstlisting}
You are a careful question-answering assistant.
Use ONLY the information in CONTEXT to answer the QUESTION.
If the answer is not in CONTEXT, respond with: "I don't know."

CONTEXT:
{retrieved_passages}

QUESTION:
{question}

Answer (one short sentence):
\end{lstlisting}

\subsection{Auto mode}
\begin{lstlisting}
Answer the QUESTION using the CONTEXT below.

CONTEXT:
{retrieved_passages}

QUESTION:
{question}

Answer:
\end{lstlisting}

\subsection{Refusal action}
The refusal action bypasses retrieval and returns a short refusal message (e.g., ``I cannot answer that.''), which is counted toward refusal rate and incurs minimal token cost.


\clearpage
\begin{thebibliography}{99}
\bibitem{lewis2020rag}
Patrick Lewis, Ethan Perez, Aleksandra Piktus, Fabio Petroni, Vladimir Karpukhin, Naman Goyal, Heinrich K{\"u}ttler, Mike Lewis, Wen-tau Yih, Tim Rockt{\"a}schel, Sebastian Riedel, and Douwe Kiela.
\newblock Retrieval-Augmented Generation for Knowledge-Intensive NLP Tasks.
\newblock In \emph{NeurIPS}, 2020. arXiv:2005.11401.

\bibitem{karpukhin2020dpr}
Vladimir Karpukhin, Barlas O{\u g}uz, Sewon Min, Patrick Lewis, Ledell Wu, Sergey Edunov, Danqi Chen, and Wen-tau Yih.
\newblock Dense Passage Retrieval for Open-Domain Question Answering.
\newblock In \emph{EMNLP}, 2020. arXiv:2004.04906.

\bibitem{guu2020realm}
Kelvin Guu, Kenton Lee, Zora Tung, Panupong Pasupat, and Ming-Wei Chang.
\newblock REALM: Retrieval-Augmented Language Model Pre-Training.
\newblock In \emph{ICML}, 2020. arXiv:2002.08909.

\bibitem{gao2024ragsurvey}
Yifan Gao, Yun Xiong, and Rui Yan.
\newblock Retrieval-Augmented Generation for Large Language Models: A Survey.
\newblock arXiv:2312.10997, 2023.

\bibitem{asai2023selfrag}
Akari Asai, Zeqiu Wu, Yizhong Wang, Aohan Zeng, and Hannaneh Hajishirzi.
\newblock Self-RAG: Learning to Retrieve, Generate, and Critique through Self-Reflection.
\newblock arXiv:2310.11511, 2023.

\bibitem{rajpurkar2018squad2}
Pranav Rajpurkar, Robin Jia, and Percy Liang.
\newblock Know What You Don't Know: Unanswerable Questions for SQuAD.
\newblock arXiv:1806.03822, 2018.

\bibitem{sutton2018rl}
Richard~S. Sutton and Andrew~G. Barto.
\newblock \emph{Reinforcement Learning: An Introduction}.
\newblock MIT Press, 2nd edition, 2018.

\bibitem{schulman2017ppo}
John Schulman, Filip Wolski, Prafulla Dhariwal, Alec Radford, and Oleg Klimov.
\newblock Proximal Policy Optimization Algorithms.
\newblock arXiv:1707.06347, 2017.

\bibitem{levine2020offline}
Sergey Levine, Aviral Kumar, George Tucker, and Justin Fu.
\newblock Offline Reinforcement Learning: Tutorial, Review, and Perspectives on Open Problems.
\newblock arXiv:2005.01643, 2020.

\bibitem{swaminathan2015crm}
Adith Swaminathan and Thorsten Joachims.
\newblock Counterfactual Risk Minimization: Learning from Logged Bandit Feedback.
\newblock In \emph{ICML}, 2015. arXiv:1502.02362.

\bibitem{dudik2011dr}
Miroslav Dud{\'i}k, John Langford, and Lihong Li.
\newblock Doubly Robust Policy Evaluation and Learning.
\newblock In \emph{ICML}, 2011.

\bibitem{beyer2016sre}
Betsy Beyer, Chris Jones, Jennifer Petoff, and Niall Murphy.
\newblock \emph{Site Reliability Engineering: How Google Runs Production Systems}.
\newblock O'Reilly Media, 2016.


\bibitem{routellm2024}
Isaac Ong, Amjad Almahairi, Vincent Wu, Wei-Lin Chiang, Tianhao Wu, Joseph E. Gonzalez, M. Waleed Kadous, and Ion Stoica.
\newblock RouteLLM: Learning to Route LLMs with Preference Data.
\newblock arXiv:2406.18665, 2024.

\bibitem{frugalgpt2023}
Lingjiao Chen, Matei Zaharia, and James Zou.
\newblock FrugalGPT: How to Use Large Language Models While Reducing Cost and Improving Performance.
\newblock arXiv:2305.05176, 2023. (Also appeared in \emph{TMLR}, 2024.)

\bibitem{robertson1995okapi}
Stephen~E. Robertson, Steve Walker, Susan Jones, Micheline Hancock-Beaulieu, and Mike Gatford.
\newblock Okapi at TREC-3.
\newblock In \emph{Proceedings of the Third Text REtrieval Conference (TREC-3)}, 1995.

\bibitem{robertson2009prf}
Stephen~E. Robertson and Hugo Zaragoza.
\newblock The Probabilistic Relevance Framework: BM25 and Beyond.
\newblock \emph{Foundations and Trends in Information Retrieval}, 3(4):333--389, 2009.
\newblock DOI: 10.1561/1500000019.

\bibitem{openai_models}
OpenAI.
\newblock Model documentation for GPT-4.1-nano and related APIs.
\newblock OpenAI Platform Documentation, accessed 2025.

\end{thebibliography}
\end{document}